\documentclass[twocolumn,10pt]{article}
 
\usepackage{amsmath}
\usepackage{graphicx}
\usepackage{xcolor}
\usepackage[utf8]{inputenc}
\usepackage{hhline}

\usepackage{lipsum}

\title{\textbf{COVID-19 Fake News Detection Using Bidirectional  Encoder  Representations  from  Transformers Based Models}}

\author{\textbf{Yuxiang Wang, Yongheng Zhang\footnote{The first two authors have equal contribution.}, Xuebo Li, Xinyao Yu\footnote{The last two authors have equal contribution.}} \\ Whiting School of Engineering, Johns Hopkins Unversity \\ ywang594@jhu.edu, yzhan470@jhu.edu, xli248@jhu.edu, xyu63@jhu.edu}
\begin{document}
\date{}
\maketitle
\begin{abstract}
    Nowadays, the development of social media allows people to access the latest news easily. During the COVID-19 pandemic, it is important for people to access the news so that they can take corresponding protective measures. However, the fake news is flooding and is a serious issue especially under the global pandemic. The misleading fake news can cause significant loss in terms of the individuals and the society. COVID-19 fake news detection has become a novel and important task in the NLP field. However, fake news always contain the correct portion and the incorrect portion. This fact increases the difficulty of the classification task. In this paper, we fine tune  the  pre-trained Bidirectional  Encoder  Representations  from  Transformers  (BERT)  model as our base model. We add BiLSTM layers and CNN layers on the top of the finetuned BERT model with frozen parameters or not frozen parameters methods respectively. The model performance evaluation results showcase that our best model (BERT finetuned model with frozen parameters plus BiLSTM layers) achieves state-of-the-art results towards COVID-19 fake news detection task. We also explore keywords evaluation methods using our best model and evaluate the model performance after removing keywords.
 %   \vspace{-15mm}
    \paragraph{Keywords:} \noindent COVID-19 Fake news, BERT, Finetune, BiLSTM, CNN 
\end{abstract}

%	\maketitle
	%do not write an abstract
	\section{Introduction}
	\subsection{Background} 
	From the past year, the whole world has gone through the COVID-19 pandemic. Twitter, Facebook, Instagram and many other social platforms update news on pandemics every day. In this project, we will use and fine tune the pre-trained Bidirectional Encoder Representations from Transformers (BERT) based models to train social media news posts, which are already known for truth or fake, for better recognizing those possible false news that may appear in the future. 

	\subsection{Related Work} Adhikari \cite{DBLP:journals/corr/abs-1904-08398} presents the first application of BERT to document classification. Aggarwal \cite{aggarwal2020classification} demonstrates classification of fake news by fine-tuning deep bidirectional transformers based language Model. Devlin \cite{DBLP:journals/corr/abs-1810-04805} introduces a new language representation model called BERT. Gundapu \cite{DBLP:journals/corr/abs-2101-00180} introduces an ensemble of three transformer models (BERT, ALBERT, and XLNET) to detect fake news. Gupta \cite{DBLP:journals/corr/abs-2101-05953} builds a model that makes use of an abusive language detector coupled with features extracted via Hindi BERT and Hindi FastText models and metadata. Kaliyar \cite{kaliyar2021fakebert} proposes a BERT-based deep learning approach by combining different parallel blocks of the single-layer deep Convolutional Neural Network having different kernel sizes and filters with the BERT. Kula\cite{kula2020application} presents a hybrid architecture connecting BERT with RNN. Liu \cite{liu2019two} treats fake news detection as fine-grained multiple-classification task and use two similar sub-models to identify different granularity labels separately.\:Pham \cite{pham2019transferring} explores encoding news title pairs and transforms into new representation space. Pham-Hong \cite{pham-hong-chokshi-2020-pgsg} uses a stack of BERT and LSTM layers to evaluate multilingual offensive language identification in social media. Safaya \cite{safaya-etal-2020-kuisail} describes approach to utilize pre-trained BERT models with Convolutional Neural Networks for sub-task of the Multilingual Offensive Language Identification shared task. Sun \cite{DBLP:journals/corr/abs-1905-05583} investigates different fine-tuning methods of BERT on text classification task and provides a general solution for BERT fine-tuning. Tang \cite{DBLP:journals/corr/abs-1910-05786} mentions keyword extraction using Attention-based Deep Learning models with BERT. Vijjali \cite{vijjali2020two} leverages a novel fact checking algorithm that retrieves the most relevant facts concerning user claims about particular COVID-19 claims, and verifies the level of“truth” in the claim by computing the textual entailment between the claim and the true facts.
	
% 	\begin{figure*}[!ht]
%     \center
%     \includegraphics[scale=0.2]{1.png}
%     \caption{Architecture for the BERT Model\cite{DBLP:journals/corr/abs-1810-04805}}
%     \end{figure*}
	\vspace{-0.7mm}
	\section{Methods}
	\subsection{Dataset}
	The COVID-19 Fake News Detection Dataset comes from the Kaggle website\footnote{https://www.kaggle.com/elvinagammed/covid19-fake-news-dataset-nlp}.\:We have the balanced training data, which contains 6,420 data entries with variable id, tweet and label. We also have balanced testing data which contains 2,140 data entries with variable id and tweet. There are three main variables in our training dataset: 'id' indicates the id number of the tweet; 'tweet' means the actual context of the tweet/post; lastly, 'label' describes whether the news is real or fake. We combine those two datasets together and randomly split data into training set (90\%) and test set (10\%) using the set seed method. 
	
	\subsection{Setup} Data pre-processing is essential for feeding the data into BERT. The pre-processing steps can be summarized as the following steps: First, we load the dataset. Second, we perform tokenization and Encoding. We use BertTokenizer and our own tokenizer to tokenize the tweets. [SEP] and [CLS] tokens need to be added at the end and beginning of every sentence. Then, we map tokens to ids. Third, we apply pad and Truncation. BERT requires that all sentences must have the same fixed length and the max length of 512 tokens per sentence. We found out that only 10 out of 8560 rows has length that is over 512. Therefore, we set up our max length to 512. Last, we use Attention Masks. The purpose of adding the masks is to not incorporate the padded tokens into the interpretation of the sentences.
	
	\subsection{Training and Evaluation}
	Our training and evaluation procedure can be summarized as the following steps.\ First, we apply the BertForSequenceClassification model.\quad Second, we fine tune the BERT model.\ Third, we add additional layers after the fine-tuned model, including CNN and Bidirectional LSTM, for both with and without freezing the parameters in the fine-tuned model.Then, we perform training, hyperparameter tuning, and testing. Last, we investigate key words that affect the authenticity of the news. The procedure can be visualized in Figure 1.
  %  \vspace{-3mm}
    \begin{figure*}[!ht]
    \center
    \includegraphics[scale=0.2]{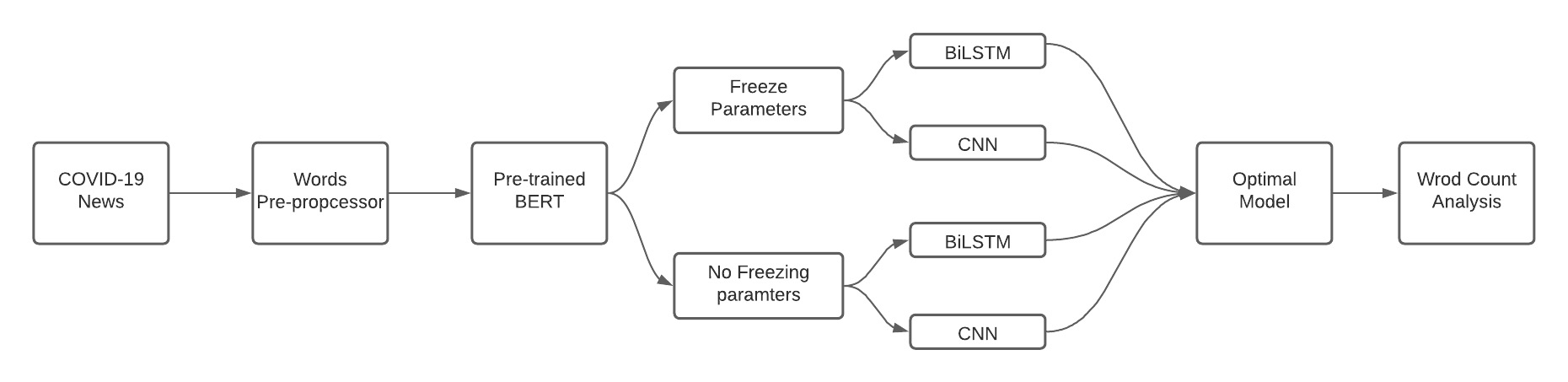}
    \caption{Pipeline}
    \end{figure*}

	\section{Model}
	We build five different models to evaluate and compare the performance of fake news classification.We define Model 1 as the BERT finetuned model. Next, we define Model 2 as the BERT finetuned model with frozen parameters plus CNN layer(s).Then, we define Model 3 as the BERT finetund model without frozen parameters plus CNN layer(s). Besides, we define Model 4 as the BERT finetuned model with frozen parameters plus BiLSTM layer(s). Lastly, we define Model 5 as the BERT finetuend model without frozen parameters plus BiLSTM layer(s).
	
	\begin{figure*}[!ht]
    \center
    \includegraphics[scale=0.5]{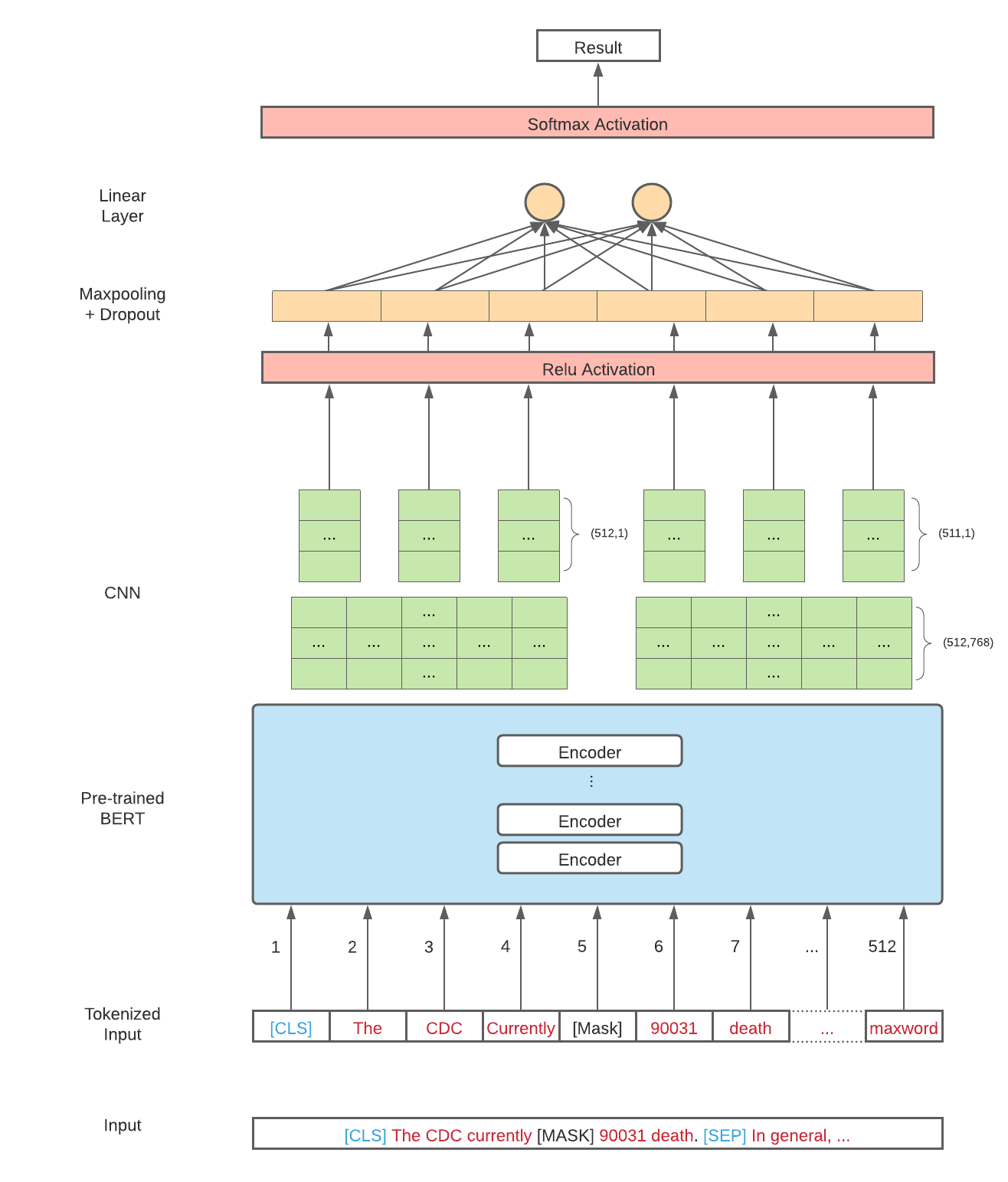}
    \caption{BERT+CNN Model Architecture}
    \end{figure*}

	\begin{figure*}[!ht]
    \center
    \includegraphics[scale=0.5]{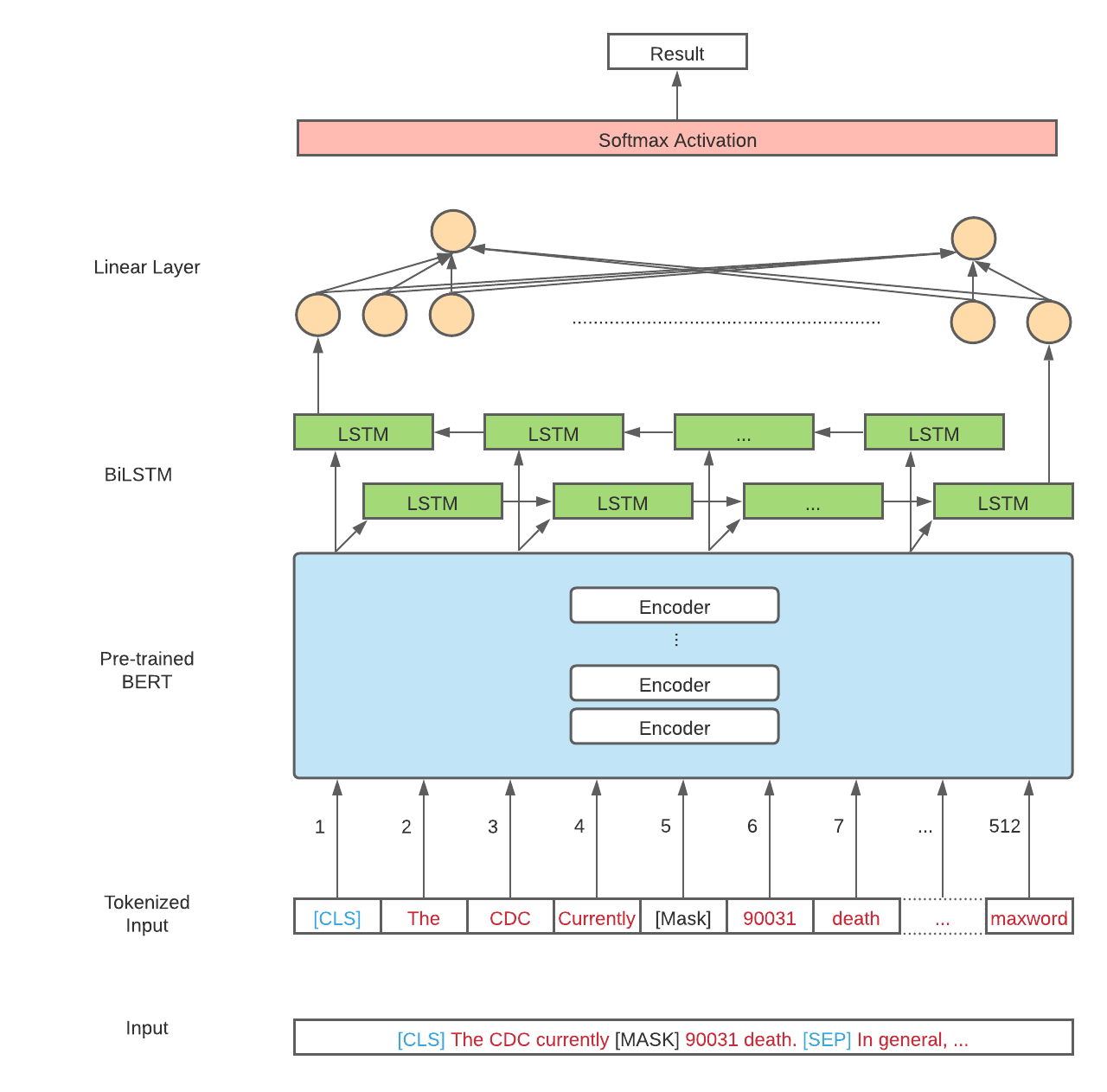}
    \caption{BERT+BiLSTM Model Architecture}
    \end{figure*}
	
	\subsection{Design Architecture}
	\paragraph{BERT finetune}For model 1, we use 2e-5 as the learning rate, and use 4 epochs to fine tune the model. %\\
%	\\
	\paragraph{BERT finetune + CNN} For model 2 and model 3, after the BERT model, we apply two convolutional layers with kernel size (1,768) and (2,768) and ReLU activation function. Then we follow a max pooling layer with the previous output size as the kernel size and the previous height of the output as the stride. After this, we add a dropout layer with the rate 0.1. Finally, we apply a linear layer with softmax activation function. The softmax activation function can be expressed as
	\begin{align*}
	    \sigma({z})_{i}=\frac{e^{z_{i}}}{\sum_{j=1}^{K} e^{z_{j}}}
	\end{align*}
	Where K = 2 because we have two classes, fake and real news. ${z}$ indicates input. The classification result is the class with the highest output of the softmax activation function. Besides, the learning rate for those two models are both 2e-5 and the epoch for those two models are both 4. The architecture can be visualized in Figure 2.
	
	\paragraph{BERT finetune + BiLSTM} For model 4 and model 5, after the BERT model, we apply 2 BiLSTM layers to model 4 and 1 BiLSTM layers to model 5. After this, we apply a linear layer with softmax activation function. Besides, the learning rate for model 4 and model 5 are both 5e-5, and the epoch used for model 4 is 10 and for model 5 is 6. The architecture can be visualized in Figure 3. 
	
% 	Compare to Figure 2, the pre-trained BERT model stays the same but the layers follow that model have changed. However, in the end, both Figure 2 and Figure 3 apply the linear layer and the softmax activation function.
	
	\section{Results}
	\subsection{Evaluation Criteria}To test our classifiers' prediction
results on fake news dataset, we use the following metrics. First, we use test accuracy as our primary metric. In our task, the test accuracy is defined as (1). 
\begin{align}
    \frac{number\hspace{1mm}of\hspace{1mm}correctly\hspace{1mm}classified\hspace{1mm}news}{total\hspace{1mm}number\hspace{1mm}of\hspace{1mm}news}
\end{align}
\\
\noindent The second matric we use is train loss. We use cross-entropy loss can be defined as (2). The cross-entropy compares the model’s prediction with the label which is the true probability distribution.
\begin{align}
    L=-(y \log (p)+(1-y) \log (1-p))
\end{align}

\noindent The third matric we use is ROC AUC score. The ROC AUC stands for the area under the curve of ROC. The range of ROC AUC score is 0 to 1, and a large auc value for a model indicates a good performance of the prediction. The last matric we use is F1 score. It ranges from 0 to 1 and is calculated from the precision and the recall of our test results. The precision is the number of true positive results divided by the number of all positive results, and the recall is the number of true positive results divided by the number of all samples that should have been identified as positive. The higher the score, the better performance it indicates. The formula of F1 score can be written as (3).\\
	\begin{align}
        F1 \hspace{1mm} score = 2 \cdot \frac{precision\cdot recall}{precision+recall}
    \end{align}
	
	    \begin{center}
    	\begin{table}[htb]
        \centering
                
    %	\textbf{Table 1} 
        \caption{Model Evaluation}
        
    	\setlength{\tabcolsep}{0.5mm}{
        \begin{tabular}{p{1.5 cm}|c|c|c|c} 
         \hline
        \quad Model & Test acc &Train loss& ROC AUC & F1 score \\
       % \hhline{=====}
        \hline
        \: Model 1 & 0.9579 & 0.0036 & 0.9586 & 0.9607 \\
     %   \hline
       \: Model 2 &
     0.9591 & 0.0200 & 0.9589 & 0.9622 \\
      %  \hline
        \: Model 3
     & 0.9439 & 0.0211 & 0.9449 & 0.9474 \\
       % \hline
        \: Model 4& 0.9614 & 0.0197 & 0.9607 & 0.9646 \\
       % \hline
        \: Model 5 & 0.9346 & 0.0227 & 0.9351 & 0.9389 \\
        \hline
      
        \end{tabular}}
        \end{table}
        \end{center}
        
	\subsection{Performance\:Analysis} \text{From the results},\:model 4  has the highest test accuracy, ROC AUC and F1 score. The performance of model 2 is better than that of the BERT fine-tuned model as well. It is in our expectation that model 4 performs the best since BiLSTM considers the context before and after the target words.
    
    \begin{figure*}[!ht]
    \center
    \includegraphics[scale=0.7]{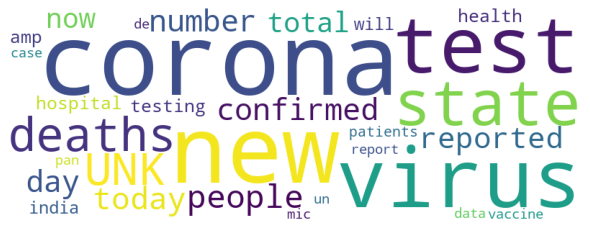}
    \caption{\label{keywords}Most Frequent Words in Our Prediction}
    \end{figure*}
    
    \subsection{Keywords in Fake News}
	\paragraph{Word count}
	We count and sort the words in sentences which are classified as fake news by our best model to obtain keywords. For example, excluding some commonly used prepositions, some of the keywords can be visualized in Figure 4.
	
	\paragraph{Frequent words and model performance}
	We delete those top frequent words listed above in our inputs, and see if the model performance changes after removing those words. As a result, the model performance does not change. This indicates that top frequent words do not usually sololy contribute to the overall performance.

	\section{Discussion}
	
	\subsection{Dataset Limitation} In terms of the size of the dataset, we could collect more fake news data so that the model can be better trained. We also could try different data set split ways to find a more reasonable one.
	
	\subsection{Model Hyperparameters} In the future, there is still space for trying different values of hyperparameters, such as learning rate and number of additional layers.

	\subsection{Model Structure} The combination of not pre-trained model with additional layers could possibly improve the performance. Pre-trained model is representative for general tasks but might not for this specific case. In addition, we could try different additional layers other than BiLSTM and CNN. One example could be GRU.
	
    \subsection{Model Evaluation} We realize that the performance of models with frozen parameters in the fine-tuned model improves, and the performance of models without frozen parameters in the fine-tuned model does not improve. The reason could be that the size of the dataset does not have enough support to learn those architectures without frozen parameters.  We also find out that adding additional layers will improve the accuracy because the new model can capture more information on the dataset. Overall, the performance of adding BiLSTM layer(s) is better than that of adding CNN layers under the same condition. We believe that the result is reasonable because BiLSTM consider the context of the sentence where CNN does not. In addition, the performances among five models are relatively similar. In order to achieve a stronger true conclusion, we need to consider use other methods such as cross validation. 
	
	\subsection{Keywords Evaluation} There are plenty of ways to find keywords that contribute to the fake news detection. For example, we can find keywords by analyzing the Attention Layer in BERT. Moreover, we can also find keywords by tracing the gradient value during the backpropagation procedure.

	\bibliographystyle{plain}
	\bibliography{references.bib}
\end{document}